\documentclass[a4paper]{esannV2}
\usepackage{graphicx}
\usepackage[latin1]{inputenc}
\usepackage{amssymb,amsmath,array}

\usepackage{lipsum}
\usepackage{xcolor}
\usepackage{hyperref}

\newcommand{\roundsymb}[1]{\text{$\left\lfloor#1\right\rceil$}}

\newcommand{\E}[1]{\text{$\mathbb{E}\left[#1\right]$}}

\newcommand{\detv}[1]{\text{$\mathit{\lowercase{#1}}$}}
\newcommand{\detvmaj}[1]{\text{$\mathit{\uppercase{#1}}$}}
\newcommand{\detvect}[1]{\text{$\boldsymbol{\lowercase{\mathrm{#1}}}$}}
\newcommand{\matr}[1]{\text{$\boldsymbol{\uppercase{\mathrm{#1}}}$}}
\newcommand{\set}[1]{\text{$\mathcal{\uppercase{#1}}$}}
\newcommand{\card}[1]{\text{$\left|#1\right|$}}

\newcommand{\Xhds}{\matr{\Xi}}
\newcommand{\xhds}[1]{\text{$\detvect{\xi}_{#1}$}}

\newcommand{\Xlds}{\matr{X}}
\newcommand{\xlds}[1]{\text{$\detvect{x}_{#1}$}}

\newcommand{\ndata}{\detvmaj{N}}
\newcommand{\dimhds}{\detvmaj{M}}
\newcommand{\dimlds}{\detvmaj{P}}
\newcommand{\nscales}{\detvmaj{H}}

\newcommand{\dhds}[2]{\text{$\detv{\delta}_{#1#2}$}}
\newcommand{\dshds}[2]{\text{$\detv{\delta}_{#1#2}^2$}}
\newcommand{\dlds}[2]{\text{$\detv{d}_{#1#2}$}}
\newcommand{\dslds}[2]{\text{$\detv{d}_{#1#2}^2$}}

\newcommand{\cf}{\detvmaj{C}}

\newcommand{\cftsne}{\text{$\cf_{t-SNE}$}}

\newcommand{\simhds}[2]{\text{$\detv{\sigma}_{#1#2}$}}
\newcommand{\simhdsss}[3]{\text{$\detv{\sigma}_{#1#2#3}$}}
\newcommand{\simhdsms}[2]{\text{$\overline{\detv{\sigma}}_{#1#2}$}}
\newcommand{\simlds}[2]{\text{$\detv{s}_{#1#2}$}}

\newcommand{\simhdst}[2]{\text{$\detv{\sigma}_{#1#2,t}$}}
\newcommand{\simldst}[2]{\text{$\detv{s}_{#1#2,t}$}}

\newcommand{\prechds}[1]{\text{$\detv{\pi}_{#1}$}}
\newcommand{\prechdsss}[2]{\text{$\detv{\pi}_{#1#2}$}}

\newcommand{\pxt}{\text{$\detvmaj{K}_{\star}$}} 
\newcommand{\pxth}[1]{\text{$\detvmaj{K}_{#1}$}} 

\newcommand{\ind}{\set{I}}
\newcommand{\indi}[1]{\text{$\ind\backslash\{#1\}$}}

\newcommand{\knhds}[2]{\text{$\detv{\nu}_{#1}^{#2}$}}
\newcommand{\knlds}[2]{\text{$\detv{n}_{#1}^{#2}$}}

\newcommand{\qnx}[1]{\text{$\detvmaj{Q}_{\text{NX}}\left(#1\right)$}} 
\newcommand{\rnx}[1]{\text{$\detvmaj{R}_{\text{NX}}\left(#1\right)$}} 
\newcommand{\rnxnoarg}{\text{$\detvmaj{R}_{\text{NX}}$}} 
\newcommand{\auc}{\text{$\text{AUC}$}}

\begin{document}
\title{Perplexity-free Parametric $t$-SNE}

\author{Francesco Crecchi$^1$, Cyril de Bodt$^2$, Michel Verleysen$^2$,\\ John A. Lee$^2$ and Davide Bacciu$^1$
%
\thanks{CdB is a FNRS Research Fellow. JAL is a FNRS Senior Research Associate.}
%
\vspace{.3cm}\\
%
1- Universit\'{a} di Pisa - Dipartimento di Informatica \\
 Largo Bruno Pontecorvo, 3, 56127 Pisa - Italy
%
\vspace{.1cm}\\
2 - Universit\'{e} catholique de Louvain - ICTEAM \\
 Place du Levant 3 L5.03.02, 1348 Louvain-la-Neuve - Belgium
}

\maketitle

\begin{abstract}
The $t$-distributed Stochastic Neighbor Embedding ($t$-SNE) algorithm is a ubiquitously employed dimensionality reduction (DR) method. 
Its non-parametric nature and impressive efficacy motivated its parametric extension. 
It is however bounded to a user-defined perplexity parameter, restricting its DR quality compared to recently developed multi-scale perplexity-free approaches. 
This paper hence proposes a multi-scale parametric $t$-SNE scheme, relieved from the perplexity tuning and with a deep neural network implementing the mapping. 
It produces reliable embeddings with out-of-sample extensions, competitive with the best perplexity adjustments in terms of neighborhood preservation on multiple data sets. 
\end{abstract}

\section{Introduction}
\label{sec:intro}

Dimensionality reduction (DR) aims at appropriately mapping high-dimensional (HD) data sets into low-dimensional (LD) spaces. 
The HD~neighborhoods preservation typically characterizes the LD embedding relevance. 
Different paradigms have been studied in DR, from linear projections as principal component analysis (PCA) to nonlinear distance preservation schemes \cite{lee2007nonlinear}. 
These being strongly limited by the norm concentration phenomenon \cite{lee2011shift}, neighbor embedding approaches as Stochastic Neighbor Embedding (SNE) \cite{hinton2002stochastic} and extensions \cite{venna2010information, bunte2012stochastic} have been developed, reaching outstanding DR performances. 
Remarkably, the $t$-SNE algorithm \cite{maaten2008visualizing} acquired tremendous popularity among the DR community. 
This motivated the design of $t$-SNE variants, such as its parametric version \cite{van2009learning}. 
Parametric schemes can indeed easily deal with very large-scale databases, by learning the projection on a random subset only and mapping the remaining data samples afterward \cite{bunte2012general}. 
They can also nicely adapt to online settings, by adding new data points to an existing embedding. 

However, $t$-SNE requires the user to choose a perplexity to tune the widths of its HD Gaussian neighborhoods. 
While such a single-scale method well retains neighborhood sizes near to the perplexity but without attaining similar performances for the other neighborhoods, multi-scale approaches typically much better recover both local and global HD structures \cite{lee2015multi}, and present the key advantage to avoid relying on a perplexity parameter. 
Nevertheless, no parametric extensions of these schemes have been proposed yet. 

This paper hence presents a parametric version of multi-scale $t$-SNE \cite{cdb2018mstsne}, generalizing parametric $t$-SNE \cite{van2009learning} in a perplexity-free method with more reliable neighborhood preservation. 
The neural network architecture implementing the HD-LD mapping in \cite{van2009learning} has been refined according to recent deep learning insights. 
Experiments highlight competitive results with parametric $t$-SNE and show more efficient management of datasets of heterogeneous structure.

This paper is structured as follows: Section~\ref{sec:sne_tsne_mstsne} first summarizes SNE, $t$-SNE and multi-scale $t$-SNE, while Section~\ref{sec:ptsne_pmstsne} reviews parametric $t$-SNE and introduces its proposed multi-scale, perplexity-free version. Section~\ref{sec:drqa} describes DR quality assessment and its application in the parametric setting. Section~\ref{sec:exp} presents the experiments, while Section~\ref{sec:concl} sketches conclusions and future works.

\section{SNE, $t$-SNE and multi-scale $t$-SNE}
\label{sec:sne_tsne_mstsne}

Let $\Xhds=\left[\xhds{i}\right]_{i=1}^{\ndata}$ be a set with $\ndata$ points in a HD~space (HDS) with dimension $\dimhds$. Let $\Xlds=\left[\xlds{i}\right]_{i=1}^{\ndata}$ model it in a $\dimlds$-dimensional space (LDS), $\dimlds\leq\dimhds$. The HD~(LD) distance between the $i^{\text{th}}$ and $j^{\text{th}}$ points is noted $\dhds{i}{j}$ ($\dlds{i}{j}$). 
SNE introduces HD~and LD~similarities, for $i\in\ind=\{1,\hdots,\ndata\}$ and $j\in\indi{i}$ \cite{hinton2002stochastic}: 
\begin{equation*}
\simhds{i}{j}=\frac{\exp\left(-\prechds{i}\dshds{i}{j}/2\right)}{\sum_{k\in\indi{i}}\exp\left(-\prechds{i}\dshds{i}{k}/2\right)}, \text{ } \simlds{i}{j}=\frac{\exp\left(-\dslds{i}{j}/2\right)}{\sum_{k\in\indi{i}}\exp\left(-\dslds{i}{k}/2\right)}, \text{ } \simhds{i}{i}=\simlds{i}{i}=0.
\end{equation*}
A binary search tunes the precision $\prechds{i}$ to set the perplexity of the distribution $\left[\simhds{i}{j};j\in\indi{i}\right]$ to a user-specified soft neighborhood size $\pxt$: $\prechds{i}$ such that $\log{\pxt} = -\sum_{j\in\indi{i}} \simhds{i}{j} \log{\simhds{i}{j}}$. SNE then optimizes the LDS by minimizing the sum of the KL divergences between the HD~and LD~similarity distributions. 

In addition to symmetrizing the similarities, $t$-SNE uses a LD Student $t$-distribution with one degree of freedom, overcoming the crowding problem \cite{maaten2008visualizing}: 
\begin{equation*} 
\simhdst{i}{j} = \frac{\simhds{i}{j}+\simhds{j}{i}}{2\ndata}, \text{ }\simldst{i}{j} = \frac{1}{\left(1+\dslds{i}{j}\right)\sum_{k\in\ind,l\in\indi{k}}\left(1+\dslds{k}{l}\right)^{-1}}, \text{ }\simldst{i}{i} = 0.
\end{equation*}
Gradient descent then minimizes the same cost function as SNE, $\cftsne = \sum_{i\in\ind,j\in\indi{i}} \simhdst{i}{j}\log\left(\simhdst{i}{j}\left/\simldst{i}{j}\right.\right)$. 
In practice, the $\pxt$ parameter is frequently arbitrarily set. 
Small values shape the LDS by mainly neglecting mid- and large-scale data interactions, due to the quickly vanishing HD similarities. 
Large perplexities lead to close to uniform HD similarities, impairing the small neighborhood reproduction. 
These difficulties are solved by multi-scale methods \cite{lee2015multi}, which preserve both local and global structures at once in the LDS by accounting for various perplexities. 
Multi-scale $t$-SNE \cite{cdb2018mstsne} refines the HD similarities as
\begin{equation*}
\simhdsss{h}{i}{j}=\frac{\exp\left(-\prechdsss{h}{i}\dshds{i}{j}/2\right)}{\sum_{k\in\indi{i}}\exp\left(-\prechdsss{h}{i}\dshds{i}{k}/2\right)}, \text{ } \simhdsms{i}{j}=\frac{1}{\nscales}\sum\nolimits_{h=1}\nolimits^{\nscales}\frac{\simhdsss{h}{i}{j}+\simhdsss{h}{j}{i}}{2\ndata}, \text{ } \simhdsss{h}{i}{i}=\simhdsms{i}{i}=0,
\end{equation*}
where $\prechdsss{h}{i}$ is fixed as in SNE using perplexity $\pxth{h}=2^h$ and $\nscales=\roundsymb{\log_2\left(\ndata\left/\pxth{1}\right.\right)}$ is the number of scales. 
The $\left[\simhdsms{i}{j}\right]_{i,j=1}^{\ndata}$ similarities are then matched with $\left[\simldst{i}{j}\right]_{i,j=1}^{\ndata}$ by optimizing SNE cost function. 
Conveniently, no more perplexity parameter is needed, the HD structure being faithfully quantified in a data-driven way. 

\section{Parametric $t$-SNE and multi-scale parametric $t$-SNE}
\label{sec:ptsne_pmstsne}

In its original formulation, $t$-SNE \cite{maaten2008visualizing} is a non-parametric manifold learner. The main limitation of the non-parametric manifold learners is that they do not provide a parametric mapping between the high-dimensional data space and the low-dimensional latent space, making it impossible to embed new data points without the need to re-train the model. The out-of-sample extension of $t$-SNE, namely parametric $t$-SNE \cite{van2009learning}, parametrizes the non-linear mapping between the data space and the latent space through a feed-forward neural network. 

As a neural network with sufficient hidden layers (with non-linear activation functions) is known to be capable of approximating arbitrarily complex non-linear functions \cite{cybenko1989approximation}, it is used here for learning the parametric mapping $f: \Xhds \rightarrow \Xlds$ from the data space $\Xhds$ to the low-dimensional latent space $\Xlds$. The weights of the neural network are learned to minimize the Kullback-Leibler (KL) divergence $\cftsne$ between the HD and LD probability distributions\footnote{Refer to \cite{van2009learning} for training procedure details.}. 
The asymmetric nature of the KL divergence leads the minimization to focus on modeling large $\simhds{i}{j}$'s by large $\simlds{i}{j}$'s. Hence the objective function focuses on modeling similar data points close together in the latent space. As a result of this, parametric $t$-SNE focuses on preserving the local structure of the data. 

The \textit{locality} of the method is retained according to the main hyper-parameter in $t$-SNE: the perplexity $\pxt$. It can be interpreted as the size of the soft Gaussian neighborhood in the HD space, and it has a strong impact on the modeled representation. The optimal tuning of this parameter is the key to obtain LD embeddings retaining most of the local structure of the data. The user can be relieved from this duty by employing a multi-scale approach \cite{cdb2018mstsne}, which automatically takes into account various perplexities to preserve both local and global structures at once. 

The main contribution of this work is to introduce the multi-scale approach into parametric $t$-SNE, leading to a parametric manifold-learner capable to retain both the local and the global structure of the data, without the need of tuning the perplexity parameter.
To achieve this, we updated the original parametric $t$-SNE neural network by using $\simhdsms{i}{j}$ to compute HD similarities, in a multi-scale fashion. Moreover, we replaced logistic activation functions with piecewise-linear ones (i.e. $ReLU$s) which do not saturate during training. This simple architectural choice allowed us to ease the training procedure by dismissing the unsupervised pre-training step introduced in the original parametric $t$-SNE paper \cite{van2009learning}. The source code is publicly available at \url{https://github.com/FrancescoCrecchi/Multiscale-Parametric-t-SNE}.

\section{Assessing the quality of dimensionality reduction}
\label{sec:drqa}

Quality criteria for unsupervised DR usually measure the HD~neighborhood preservation in the LDS~\cite{lee2009quality, venna2010information}. 
The $K$ nearest neighbor sets of $\xhds{i}$ and $\xlds{i}$ in the HDS~and LDS are noted $\knhds{i}{K}$ and $\knlds{i}{K}$. 
Their average normalized agreement is $\qnx{K} = \sum_{i\in\ind} \card{\knhds{i}{K} \cap \knlds{i}{K}}\left/\left(K\ndata\right)\right.\in\left[0,1\right]$. 
As random LD~points yield $\E{\qnx{K}}=K/\left(\ndata-1\right)$, distinct neighborhood sizes are confronted using $\rnx{K} = \left(\left(\ndata-1\right)\qnx{K}-K\right)\left/\left(\ndata - 1 - K\right)\right.$, displayed with a log-scale for $K$ to favor the typically prevailing closer neighbors \cite{lee2013type}. 
The area under this curve, $\auc = \left(\sum_{K=1}^{\ndata-2}\rnx{K}/K\right)\left/\left(\sum_{K=1}^{\ndata-2}K^{-1}\right)\right.\in \left[-1,1\right]$ 
increases with DR quality, assessed at all scales with an emphasis on smaller ones \cite{lee2015multi}. 

In the context of parametric mappings, $\rnxnoarg$ curves are computed on both the training set and the union of the training and test sets in Section~\ref{sec:exp}, i.e. before and after projecting the test samples. 
Parametric maps are indeed commonly used to add new points to an existing LDS. 
Thus, after the test samples projection, one should assess the quality of the whole augmented LDS and not only of the test set, as the data neighborhoods involve both training and test points. 

\section{Experiments}
\label{sec:exp}

\begin{figure}
    \hspace*{-1cm}
    \setlength{\tabcolsep}{0pt} 
    \centering
    \begin{tabular}{m{.05\textwidth}m{.5\textwidth}m{.5\textwidth}}
        \rotatebox{90}{\textbf{Mnist}} & \includegraphics[width=.5\textwidth]{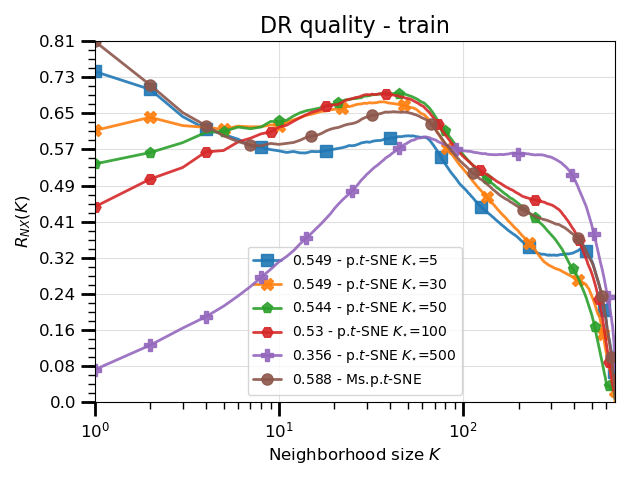} & \includegraphics[width=.5\textwidth]{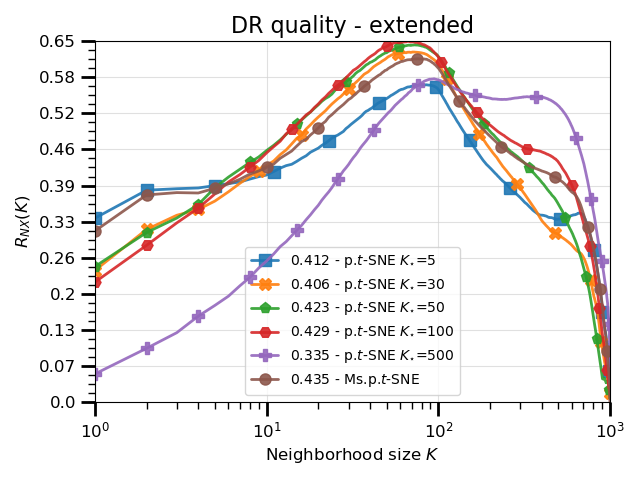} \\
        \rotatebox{90}{\textbf{Coil-20}} & \includegraphics[width=.5\textwidth]{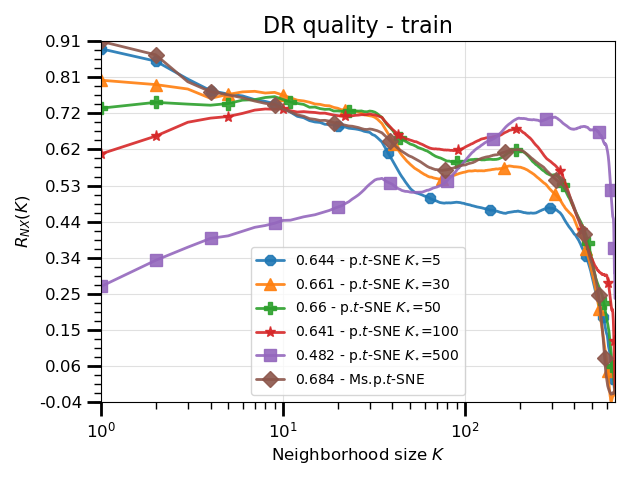} & \includegraphics[width=.5\textwidth]{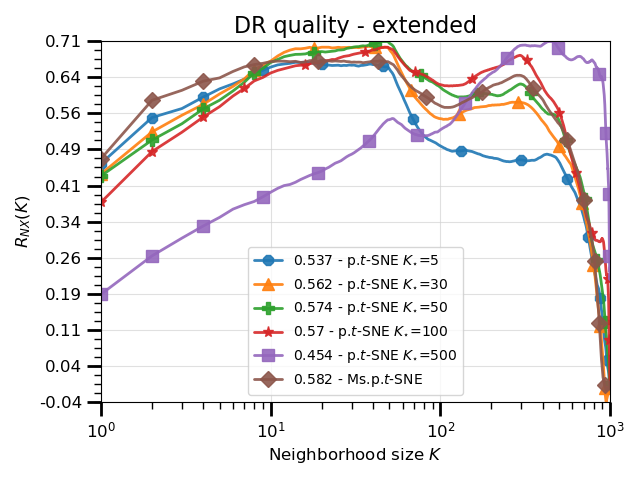} \\
        \rotatebox{90}{\textbf{Ecoli}} & \includegraphics[width=.5\textwidth]{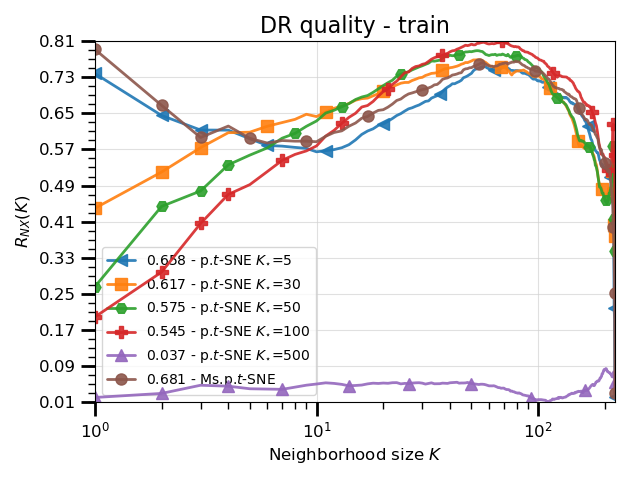} & \includegraphics[width=.5\textwidth]{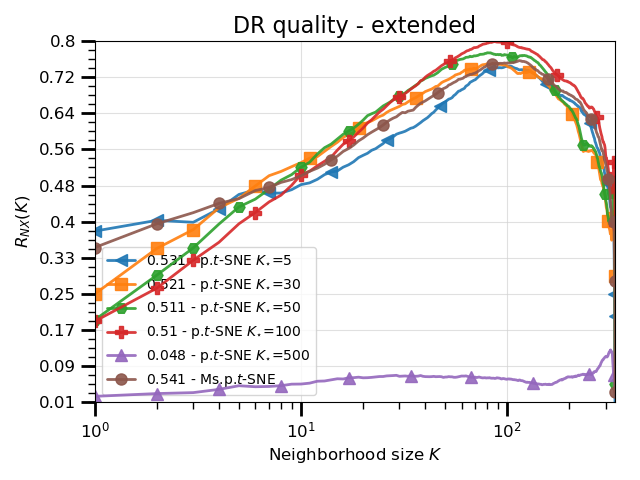} \\
        \rotatebox{90}{\textbf{Helix}} & \includegraphics[width=.5\textwidth]{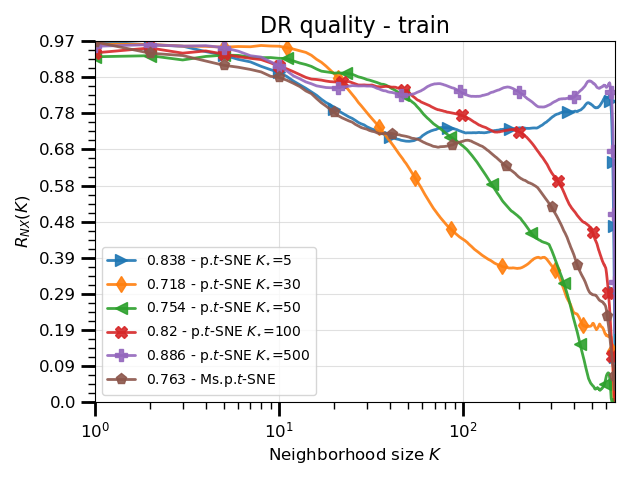} & \includegraphics[width=.5\textwidth]{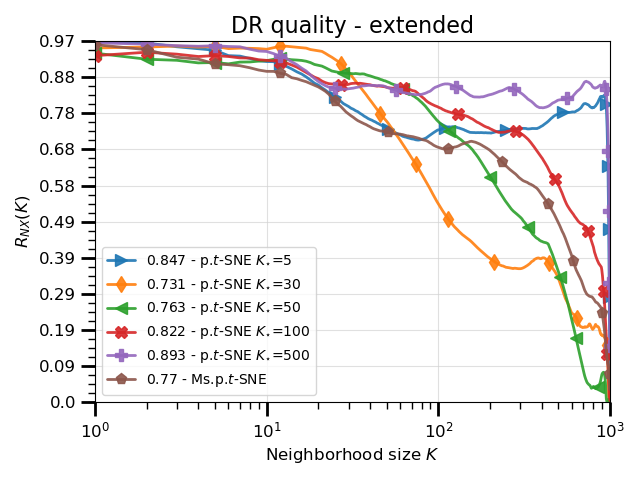} 
    \end{tabular}
    \caption{Quality curves for all data sets and all compared methods for training and extended scenarios. Each curve reports $\rnxnoarg$. 
    The higher the curve, the better.
    The corresponding $\auc$ value for each method is reported in the legend, right before the name.
    }
    \label{fig:experiments}
\end{figure}

We compared the generalization capability of parametric $t$-SNE (p.$t$-SNE) for various perplexities and multi-scale parametric $t$-SNE (Ms.p.$t$-SNE) in terms of DR quality, for multiple data sets, measured by the $\rnxnoarg$ and $\auc$ metrics. We measured the model performances in the so-called \textit{extended} scenario which comprises mapping new points into an existing LDS. For a given data set, we held-out a separate test set ($30\%$ of the data). The number of hidden layers is set to four for all datasets, matching the original $t$-SNE paper NN architecture, and a grid-search is performed to choose the number of neurons per layer. Each model is then trained on training samples and performances are measured on the set obtained by merging training and test splits, as mentioned in Section~\ref{sec:drqa}. The used data sets are a subset of MNIST (N = 1000, M = 784) \cite{lecun1998christopher}, COIL-20 (N = 1440, M = 16384) \cite{nene1996columbia}, ECOLI (N = 336, M = 8) \cite{Dua:2019} and Helix data set (N = 1000, M = 3) \cite{lee2015multi}. Target dimensionality $P$ is 2 for all data sets.
Figure \ref{fig:experiments} reports the quality curves $\rnxnoarg$ for all data sets and all the compared methods. The proposed multi-scale approach outperforms its single-scale variants in almost all data sets for the training set and performs at least as well as the best single-scale variant on the extended set. More importantly, the curves show that setting the wrong perplexity value has a severe impact on the produced mapping, confirming the usefulness of the multi-scale approach as it reliably approximates the best possible perplexity choice. 

\section{Conclusions, limitations and future works}
\label{sec:concl}

In this paper, we combined the benefits of the multi-scale approaches with the parametric $t$-SNE method, to obtain a perplexity-free parametric scheme that reliably approximates both local and global structures of the data manifold without the need of tuning the perplexity hyper-parameter. This is demonstrated in the experimental assessment: the proposed multi-scale method approaches, and sometimes outperforms, the best single-scale variant on all data sets in projecting new data points into an existing embedding (\textit{extended} scenario). 

When embedding new points, all compared methods better retain global structures than local ones. The performance gap between the training and the extended scenarios for small neighborhood sizes may be due to the small sizes of our training data sets \cite{van2009learning}. 
In future works, 
we intend to investigate this phenomenon by using larger data sets and other parametric manifold learning methods (e.g. PCA, NCA, Autoencoders). Moreover,
we plan to mitigate such phenomenon by introducing a regularization term in the loss to better retain the local structure of the data when generalizing to new data points. 

\begin{footnotesize}

\bibliographystyle{unsrt}
\bibliography{references}

\end{footnotesize}


\end{document}